\documentclass{article}

\usepackage{microtype}
\usepackage{graphicx}
\usepackage{subcaption}
\usepackage{booktabs} 
\usepackage{tabularx}
\usepackage{caption}
\PassOptionsToPackage{hyphens}{url}
\usepackage{hyperref}

\usepackage{hyperref}
\usepackage{xspace} 
\usepackage[dvipsnames]{xcolor}

\newcommand{\qwenicon}{\includegraphics[height=1.8ex]{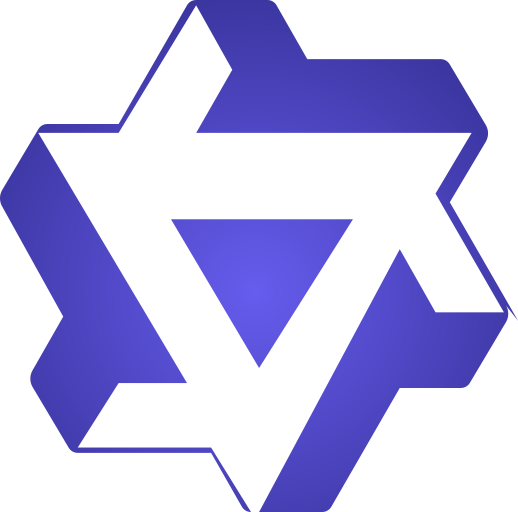}}
\newcommand{\gpticon}{\includegraphics[height=1.6ex]{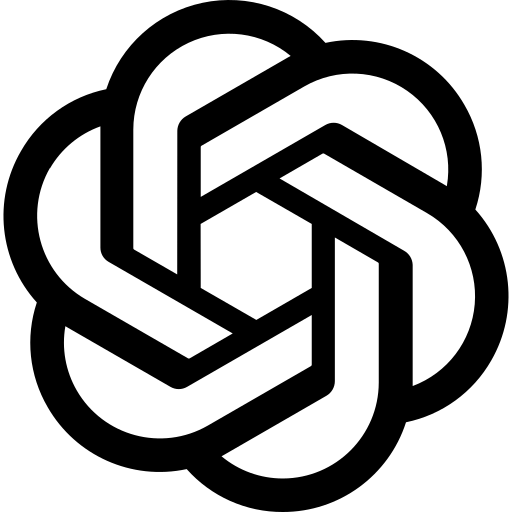}}

\usepackage[accepted]{icml2026}

\usepackage{amsmath}
\usepackage{amssymb}
\usepackage{mathtools}
\usepackage{amsthm}
\usepackage{enumitem}
\usepackage{tcolorbox}
\tcbuselibrary{breakable}

\usepackage[capitalize,noabbrev]{cleveref}

\theoremstyle{plain}

\theoremstyle{definition}

\theoremstyle{remark}

\usepackage[textsize=tiny]{todonotes}

\icmltitlerunning{Evaluating Agentic Configuration Repair for Computer Networks}

\begin{document}

\twocolumn[
  \icmltitle{Evaluating Agentic Configuration Repair for Computer Networks}



  \icmlsetsymbol{equal}{*}

  \begin{icmlauthorlist}
    \icmlauthor{Rufat Asadli}{yyy}
    \icmlauthor{Benjamin Hoffman}{equal,yyy}
    \icmlauthor{Ioannis Protogeros}{equal,yyy}
    \icmlauthor{Laurent Vanbever}{yyy}
  \end{icmlauthorlist}

  \icmlaffiliation{yyy}{Department of Information Technology and Electrical Engineering, ETH Zurich}

  \icmlcorrespondingauthor{Rufat Asadli}{rasadli@ethz.ch}

  \icmlkeywords{Machine Learning, ICML}

  \vskip 0.3in
]



\printAffiliationsAndNotice{\icmlEqualContribution}

\begin{abstract}
   Misconfigurations in computer networks remain a major source of critical Internet outages. Research is turning to Large Language Models (LLMs) to automate the complex, error-prone task of network configuration. However, even state-of-the-art models fail to resolve misconfigurations in large-scale, complex scenarios and often introduce new errors. In this work, we benchmark open- and closed-source LLMs augmented with formal network verification and context retrieval tools. We demonstrate that agentic architectures outperform base LLMs in \textit{repair efficacy} (by 12\% on average) and \textit{safety} (by 17\% on average), enabled by the ability to dynamically manage context and iteratively validate configuration repairs. 
\end{abstract}

\section{Introduction}
Network misconfigurations are a leading source of Internet outages, causing a large share of downtime in enterprise and ISP networks~\cite{janardhan2021facebook, prince2025cloudflare}.
Thus, ensuring and maintaining network correctness has been a long-standing research goal, which remains largely unrealized due to the growing complexity of network infrastructure. Formal verification and synthesis have made significant strides, but remain inapplicable to many real-world, open-ended deployments due to limited and sometimes inaccurate modeling of complex network behavior~\cite{krentsel-modelfree}. As a result, human errors continue to be the root cause of numerous impactful and costly network outages.

\paragraph{LLMs for network operations} Following their impressive software engineering performance~\cite{jimenez2024swebenchlanguagemodelsresolve, jain2024livecodebenchholisticcontaminationfree, vero2025baxbenchllmsgeneratecorrect}, LLMs stand as a more flexible alternative for automating network operations. To this end, hyperscalers have already started deploying them in production \textemdash~including ByteDance's NetAssistant~\cite{netassistant}, Alibaba's BiAn~\cite{bian}, and Meta's Confucius~\cite{confucius}. However, adoption remains limited by safety concerns: hallucinations and erroneous outputs can cause catastrophic outages.



\paragraph{Existing benchmarks} Recent works have begun to rigorously quantify the viability of LLMs in network operations. \textsc{NetArena}~\cite{netarena} provides a dynamic evaluation environment for LLMs spanning basic network analysis, verification, and configuration tasks. \textsc{NIKA}~\cite{nika} proposes a structured methodology to use LLM agents in network troubleshooting; however, it does not support the evaluation of proposed fixes for network faults. More recently, \textsc{Cornetto}~\cite{cornetto} evaluates LLM-based end-to-end configuration repair across diverse protocols and topologies at scale.

Collectively, these efforts reveal that even frontier models struggle with complex network tasks under \textit{monolithic} (i.e., single-turn) prompting. Three recurring challenges emerge: \textit{(i)} configurations span thousands of lines where noise obscures relevant signals; \textit{(ii)} single-shot attempts hinder error correction, causing models to stop at partial diagnoses; and \textit{(iii)} the semantic gap between edits and their emergent \textit{forwarding behavior} (i.e., how traffic flows through a network) makes their network-wide impact difficult to predict.
    

\paragraph{This work: Agentic repair for misconfigurations} Agentic architectures show promise to address each of these limitations \textemdash~for instance, agent-computer interfaces have proven important in LLM-driven software engineering~\cite{yang2024sweagentagentcomputerinterfacesenable, wang2025openhandsopenplatformai}.
We apply this paradigm for network configuration, with tools for dynamic context retrieval to help filter noise; iterative editing to address single-shot failures; and verifier access allowing models to safely examine network-wide reconfiguration effects without live deployment. Unlike prior agent benchmarks for networking, which focus on short diagnostics or single-command tasks~\cite{nika, netarena}, this setup targets end-to-end iterative configuration repair.

 Building on this, we evaluate open- and closed-source LLMs on \textsc{Cornetto}, equipping them with (1) dynamic context retrieval, (2) iterative search-and-replace editing, and (3) formal verification as environment feedback.

\paragraph{Key contributions} Our summarized contributions are:
\begin{itemize}
    \item We propose a minimal agentic environment for network configuration with tool calls for dynamic context retrieval, iterative repair, and verification feedback.
    \item Using our setup, we find that the agentic system noticeably improves over monolithic approaches, with 12\% gain in restoring correct network behavior and 17\% reduction in safety regressions \textit{on average}.

    \item We show that dynamic context management and continued access to a verifier positively affect the agent's safety, while the effect on efficacy is more nuanced.

\end{itemize}

\begin{figure}[t]
    \centering
    \includegraphics[width=1.0\linewidth]{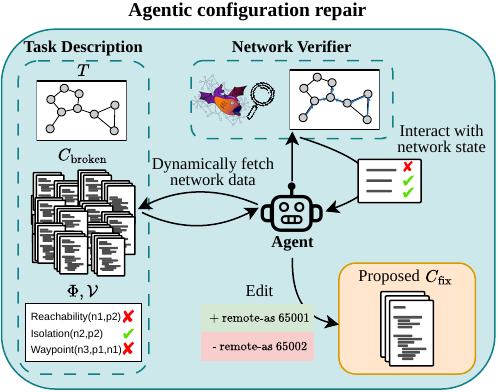}
\caption{Overview of the agentic configuration repair pipeline. Given a task description consisting of a network topology $T$, broken configurations $C_{\text{broken}}$, and violated specifications $\mathcal{V}$, the agent dynamically retrieves relevant context of its own choice, proposes search-and-replace edits, and interacts with a verifier to see the state of unresolved specifications before submitting the final $C_{\text{fix}}$.}    \label{fig:placeholder}
\end{figure}




\section{Problem Setting and Benchmark}

\begin{figure*}[h]
    \centering
    \begin{subfigure}[h]{0.49\textwidth}
        \centering
        \includegraphics[width=\textwidth]{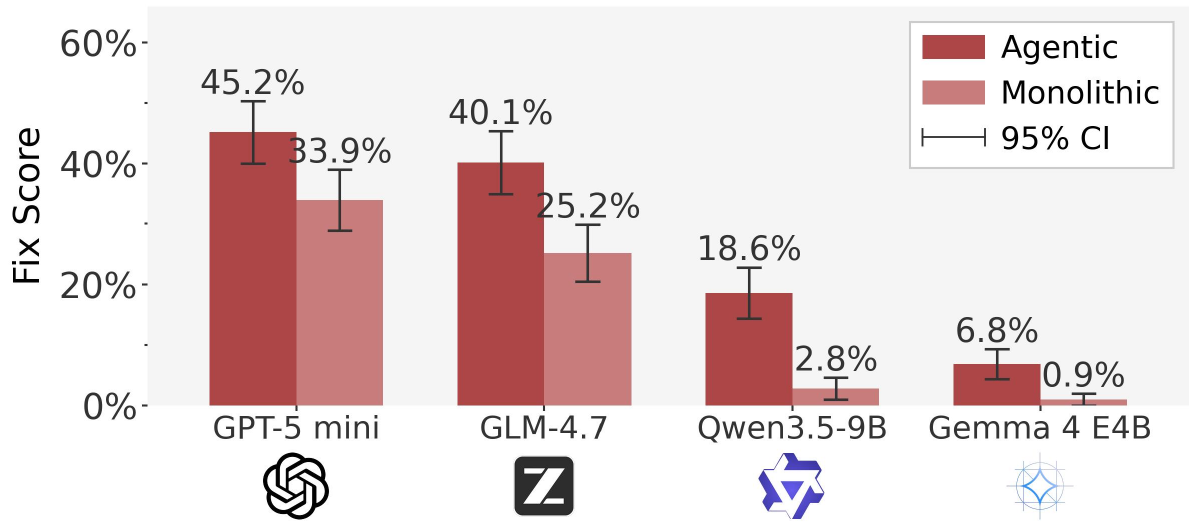}
        \caption{Fix Score ($\uparrow$ \textit{is better})}
        \label{fig:baseline_vs_agentic_fix_score}
    \end{subfigure}%
    \hfill%
    \begin{subfigure}[h]{0.49\textwidth}
        \centering
        \includegraphics[width=\textwidth]{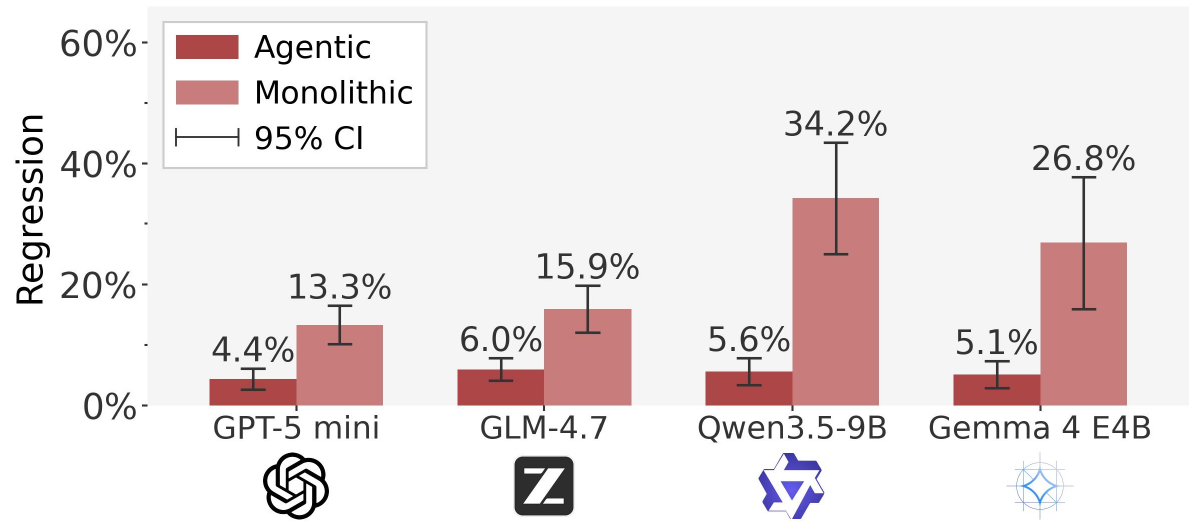}
        \caption{Regression Rate ($\downarrow$ \textit{is better})}
        \label{fig:baseline_vs_agentic_reg_rate}
    \end{subfigure}%
    \caption{Performance of the agentic pipeline (with context retrieval, without iterative verification) against monolithic baselines (in \%).}
        \label{fig:baseline_v_agentic}
\end{figure*}

\paragraph{Task definition}

\newcommand{\config}[1]{C_\text{#1}\xspace}
\newcommand{\router}[0]{\texttt{r}\xspace}
\newcommand{\prefix}[0]{\texttt{p}\xspace}

We evaluate agents on \textsc{Cornetto}~\cite{cornetto}, a benchmark 
for LLM-driven network configuration repair. 
Computer networks are governed by \textit{\textbf{network-wide configurations}}: sets of distributed, per-router text files that specify how each router processes and forwards traffic (e.g., via protocols such as BGP and OSPF). Misconfigurations in these files can violate intended network behavior (e.g., dropping or misrouting traffic), and are notoriously hard to resolve~\cite{understanding-misconfigurations}.

The benchmark defines a misconfiguration scenario as a tuple $(T, \config{gold}, \config{broken}, \Phi)$, where 
$T$ is the network topology, $\config{gold}$ is the correct 
``golden'' configuration, $\config{broken} = f(\config{gold})$ is a 
faulty configuration derived by applying a fault function $f$, and 
$\Phi$ is the set of specifications satisfied by 
$\config{gold}$ (denoted as $\config{gold} \models \Phi$). A \textit{\textbf{network specification}} $\phi \in \Phi$ captures the network's intended functionality as a boolean predicate over forwarding properties (e.g., \textit{``Router A must be able to reach prefix 10.0.0.0/8''}, or \textit{``Router B's traffic to the same prefix must pass through router C''}). 
The set of violated specifications is
$\mathcal{V} = \{\phi \in \Phi \mid \config{broken} \not\models \phi\}.$
Given the task description $\mathcal{I} = (T, \config{broken}, \mathcal{V})$, the system under test must repair the configuration by producing a configuration $\config{fix} = 
\mathcal{R}(\mathcal{I})$ such that $\config{fix} \models \Phi$.

\paragraph{Dataset}

The benchmark comprises 231 misconfiguration scenarios across real-world topologies from Topology Zoo~\cite{topologyzoo}, spanning 27 fault types (App.~\ref{app:faults}) with up to 8 simultaneous faults per scenario. Networks range up to 754 nodes and 200K configuration lines.
Table~\ref{tab:dataset} (App.~\ref{app:dataset-stats}) summarizes key dataset statistics. A 
defining characteristic is the disproportion 
between configuration perturbation and disruption in network behavior: faults 
touch less than 1\% of configuration lines on average, yet disrupt 
up to 50\% of network specifications. Thus, resolving a scenario 
requires navigating tens of thousands of lines within distributed 
configuration files to locate a handful of relevant lines, making 
precise context management a core challenge.


\paragraph{Evaluation method}
\label{sec:eval-method}

Proposed fixes are evaluated by simulating the forwarding behavior of $\config{fix}$ and checking it against the ground-truth specification set $\Phi$~\cite{cornetto}.~\footnote{The network verifier Batfish~\cite{batfish} is used along with Config2Spec~\cite{config2spec} 
to simulate network behavior and extract sets of satisfied and ground-truth specifications.} The performance for \textit{efficacy} and \textit{safety} is quantified via two 
primary metrics ($\in \left[0,1 \right]$), respectively

\[
\text{Fix Score} = \frac{|\Phi_{\text{fixed}}|}{|\Phi_{\text{fixed}}| 
+ |\Phi_{\text{unfixed}}| + |\Phi_{\text{regressed}}|},
\]
\[
\text{Regression} = \frac{|\Phi_{\text{regressed}}|}{|\Phi_{\text{fixed}}| 
+ |\Phi_{\text{unfixed}}| + |\Phi_{\text{regressed}}|},
\]

\noindent where $\Phi_{\text{fixed}}$ are initially violated 
specifications restored by the fix, $\Phi_{\text{regressed}}$ are 
previously healthy specifications broken by the fix, and 
$\Phi_{\text{unfixed}}$ are violations that remain unresolved. A 
repair is considered strictly successful only if it achieves a 
perfect fix score with zero regressions. This evaluation mirrors test-driven development in software engineering~\cite{jimenez2024swebenchlanguagemodelsresolve} where the specification set $\Phi$ 
is a collection of unit tests that can be fixed or regressed.






\section{Agentic Setup}
\label{sec:agentic-setup}

The limitations of monolithic prompting motivate three design considerations for an agentic network configuration repair system: (1) selective retrieval of relevant configuration context based on the problem description, (2) iterative editing with rollback to correct prior mistakes, and (3) interaction with a verifier to validate intermediate repairs.

We realize these design goals by developing a ReAct~\cite{yao2023reactsynergizingreasoningacting}-style agent equipped with custom tools for selective configuration retrieval, iterative editing, and verification feedback via Batfish~\cite{batfish}. Descriptions of these custom-built tools are available in App.~\ref{app:tools-descriptions}.
Given an input $\mathcal{I}=(T, \config{broken}, \mathcal{V})$, the agent is not provided with any task information initially. Instead, it retrieves the network topology $T$, the set of violated specifications $\mathcal{V}$, and the configuration files $\config{broken}$
\textit{on demand} (App.~\ref{app:agentic-input-prompt}). At each step, the agent reasons about the current state, selects a tool action, and observes the result. The agent is given a budget of $N$ steps; if no solution is submitted by then, the existing work is submitted instead~\cite {yang2024sweagentagentcomputerinterfacesenable}.

\paragraph{Dynamic context management} Feeding the full network-wide configuration, while technically possible with current context limits, is shown to be ineffective at scale~\cite{cornetto, zhang2026agenticcontextengineeringevolving}. Instead, we allow the agent to list existing routers, retrieve individual configuration files, and inspect the network topology, guided by the violated specifications describing the undesired network behaviour.

\paragraph{Iterative repair} The agent proposes configuration edits using a dedicated tool for code-patch application. The edits are performed on a working copy of $\config{broken}$, which the agent can inspect after each modification. This allows the agent to iteratively address faults, diagnosing and repairing each issue before moving to the next, rather than in a single pass.


\paragraph{Verification as feedback} Formal verification provides a key source of rigour in LLM-driven software engineering workflows~\cite{zeng2026veriequivbenchequivalencescoregroundtruthfree}. Similarly, at any point during the repair process, we allow the agent to invoke the \textsc{Cornetto} verification pipeline as a tool. It runs in the current configuration state and returns the same three-fold specification decomposition used for the final evaluation (Sec.~\ref{sec:eval-method}), enabling the agent to diagnose remaining conflicts, guide subsequent edits, or decide when to submit a solution.

\section{Preliminary Evaluation}
\label{sec:eval-results}

We first demonstrate and compare the performance of our agentic approach against monolithic LLMs in \textsc{Cornetto} and subsequently analyze the effects of our different design choices on both agent performance and behavior. 

\paragraph{Main results}
We evaluate four open- and closed-source LLMs (\textsc{GPT-5 mini}, \textsc{GLM-4.7}, \textsc{Qwen3.5-9B}, \textsc{Gemma 4 E4B}; details in App.~\ref{app:model-selection}) on all 231 misconfiguration
scenarios using our agentic setup. To ensure a fair comparison to the monolithic baseline, we initially \textit{disable} verification feedback, using it only at submission for scoring. All other tools are available with a budget of 30 iterations per task.

Our agentic approach consistently outperforms monolithic LLMs in both fix score and regression rate (Fig.~\ref{fig:baseline_v_agentic}). The benefit is especially pronounced for open-source models, which perform poorly in the monolithic setting but see their fix scores increase by up to $7\times$ under the agentic framework, while their regression rates drop from as high as 34.2\% to under 6\%. This suggests that agentic scaffolding disproportionately compensates for weaker base-model capability, while also stabilizing their safety behavior.

\begin{table}[!h]
\centering
\caption{Effects of two key design choices in the agentic pipeline: (1) verifier feedback and (2) context retrieval tools.}
\label{tab:ablations}

\small
\begin{tabular}{llccc}
\toprule
\textbf{Component} & \textbf{Setting} & \textbf{Fix} $\uparrow$ & \textbf{Regr.} $\downarrow$ \\
\midrule
\multicolumn{4}{l}{\textit{Verifier Feedback}} \\
\addlinespace[2pt]
\gpticon~\textsc{GPT-5 mini}
& Without feedback   & \textbf{45.2} & 4.4 \\
& With feedback  & 40.8 & \textbf{1.3} \\
& $\Delta$                  & \textcolor{red}{-4.4} & \textcolor{ForestGreen}{-3.1} \\
\addlinespace[3pt]
\qwenicon~\textsc{Qwen3.5-9B}
& Without feedback   & 18.6 & 5.6 \\
& With feedback  & \textbf{20.0} & \textbf{2.9} \\
& $\Delta$                  & \textcolor{ForestGreen}{+1.4} & \textcolor{ForestGreen}{-2.7} \\
\midrule
\multicolumn{4}{l}{\textit{Context Retrieval}} \\
\addlinespace[2pt]
\gpticon~\textsc{GPT-5 mini}
& Prefilled retrieval   & \textbf{49.1} & 1.3 \\
& Dynamic retrieval    & 40.8 & 1.3 \\
& $\Delta$                    & \textcolor{red}{-8.3} & \textcolor{black}{0.0} \\
\addlinespace[3pt]
\qwenicon~\textsc{Qwen3.5-9B}
& Prefilled retrieval   & 14.8 & 8.3 \\
& Dynamic retrieval    & \textbf{20.0} & \textbf{2.9} \\
& $\Delta$                    & \textcolor{ForestGreen}{+5.2} & \textcolor{ForestGreen}{-5.4} \\
\bottomrule
\end{tabular}
\end{table}

\paragraph{Design effects} We isolate two design choices, access to iterative verification feedback and dynamic context retrieval, ablating each independently (Tab.~\ref{tab:ablations}).
For these experiments, we focus on \textsc{GPT-5 mini} and \textsc{Qwen3.5-9B}, given they are the best performing closed- and open-source models from our main evaluation, respectively.
First, \textit{enabling} iterative verifier feedback benefits \textsc{Qwen3.5-9B}, achieving higher fix scores and lower regressions than the agentic setup without feedback. For \textsc{GPT-5 mini}, it cuts regressions, but also decreases the fix score, suggesting that additional verification calls make the model more conservative in its edits.

\begin{figure}[h]
    \centering
    \includegraphics[width=1.0\linewidth]{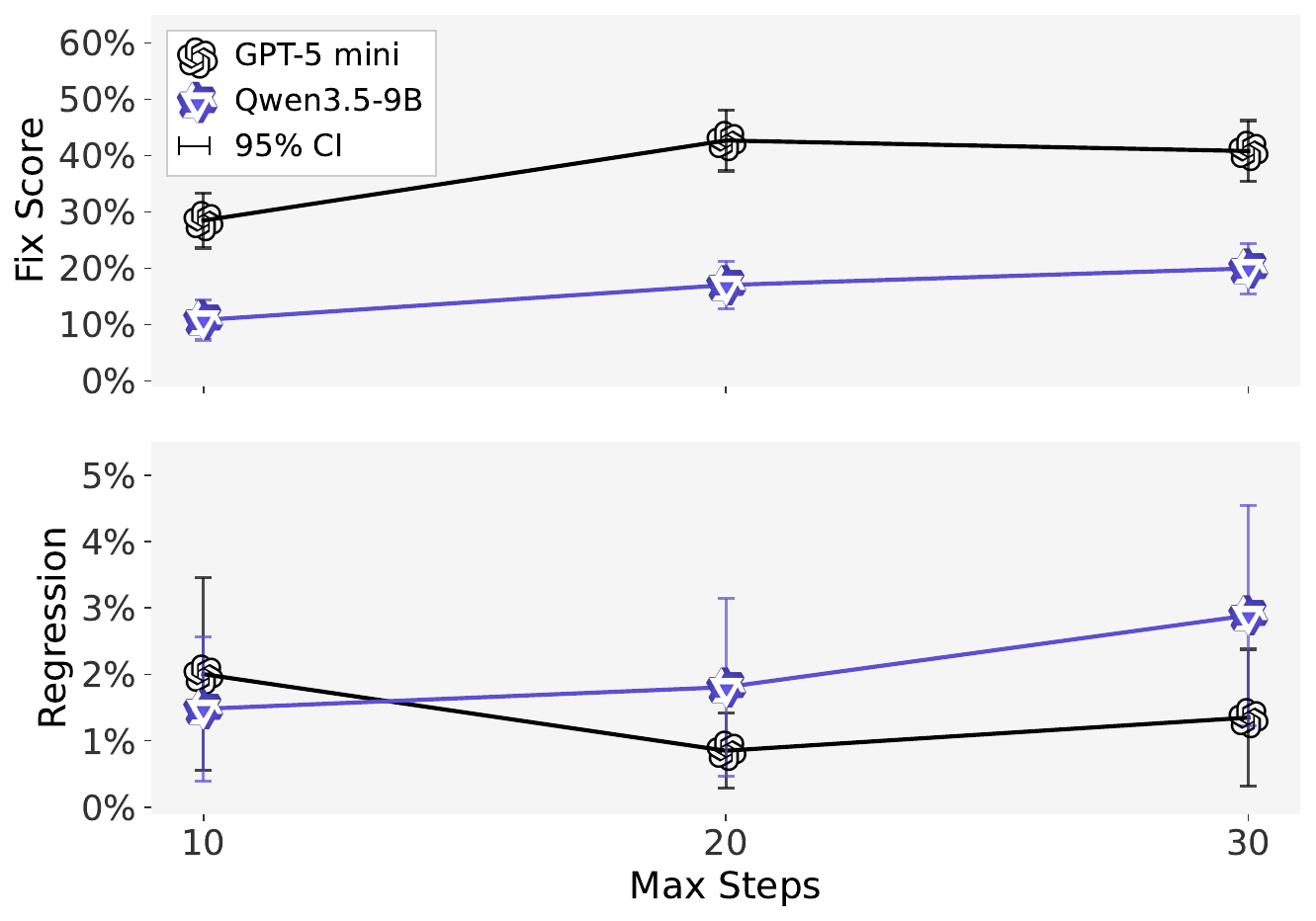}
    \caption{Effects of different maximum step budgets on agents ($\uparrow$ \textit{is better} for Fix Score,  $\downarrow$ \textit{is better} for Regression).}\label{fig:ablation_max_steps}
\end{figure}

Next, while iterative verifier feedback is enabled, we test the impact of context retrieval. \textsc{GPT-5 mini} performs best at $49\%$ fix score
when the entire problem description $(T, \config{broken}, \mathcal{V})$ is \textit{prefilled} in its context window a priori rather than \textit{dynamically} retrieved. We attribute this to the model's strong long-context capabilities, which allow it to effectively process the full scenario in a single pass, reducing the benefit of spending steps on context retrieval versus spending them on verifier feedback and editing. \textsc{Qwen3.5-9B}, despite its nominally large context window, does not benefit from prefilled inputs, as smaller models often struggle to extract relevant signals from large contexts~\cite{zhang2026agenticcontextengineeringevolving}. Instead, with targeted context retrieval, it exhibits a higher fix score ($20\%$) and lower regression ($3\%$). 

\begin{figure}[h]
    \centering
    \includegraphics[width=1.0\linewidth]{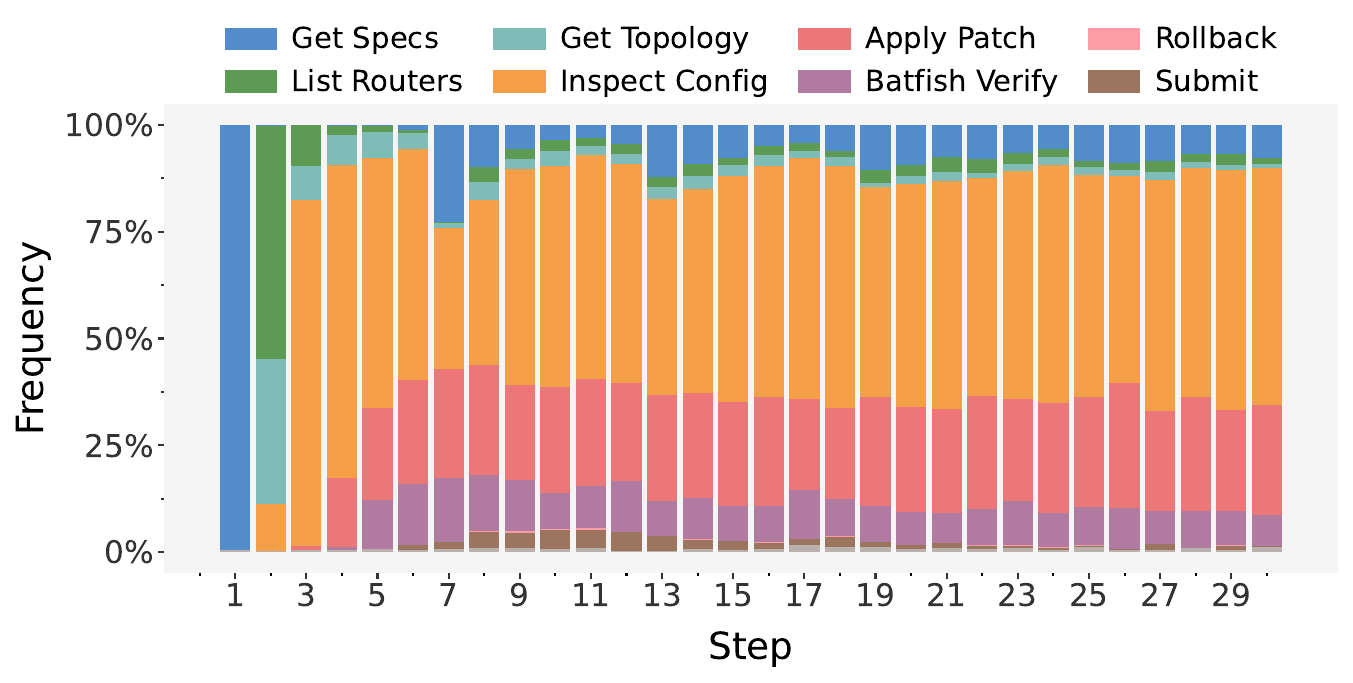}
    \caption{The frequency of the tools called at each iteration across all models over all task instances per model.}    \label{fig:tool_calling_counts_avg}
\end{figure}

Furthermore, with all tools available, we evaluate agent performance across different maximum step budgets (Fig.~\ref{fig:ablation_max_steps}). While more iterations generally improve fix scores, the effect on regressions is more nuanced: \textsc{Qwen3.5-9B}'s regression rate actually rises with longer budgets, likely reflecting the model's difficulty in retaining coherence over long \textit{interaction histories} (App.~\ref{app:pipeline-configs}).
Beyond efficacy, this analysis informs the cost profile of the agentic pipeline \textemdash~compute for self-hosted open-source models and tokens for metered proprietary ones. Since multi-turn inference is inherently more expensive than single-turn prompting, identifying the step budget at which fix scores saturate is essential for deploying agentic repair at reasonable cost (App.~\ref{app:cost-analysis}).


\paragraph{Agent trajectory} To better understand an agent's decision-making process, we evaluate their tool-calling behavior across 30 steps (Fig.~\ref{fig:tool_calling_counts_avg}), revealing a structured workflow: agents consistently begin by gathering specifications and inspecting the network topology, before transitioning to the patch application and verification loop in the middle steps.
Per-model trajectories (App.~\ref{app:per-model-trajectory}) further show that models tend to exhaust their budget rather than converging early.
Future work could explore tool associations and more granular feedback to better understand agent decision-making.
\section{Limitations and Future Work}
We show the benefits of our agentic approach over base LLMs for network configuration repair, improving efficacy and safety. However, the interactions among design choices introduce trade-offs that necessitate further investigation:

\paragraph{LLM-network interfaces} Unlike offline verification, richer interfaces, such as live routing tables and packet traces, would expand coverage to transient and performance-related faults that static analysis alone cannot diagnose.

\paragraph{Safety guardrails} 
More proactive guardrails to pre-screen edits against known invariants, along with human-in-the-loop approval for high-risk changes, could further improve the viability of agentic repair for real-world deployments.

\paragraph{Fine-tuning} Iterative decision-making in agentic systems also impacts training methodologies: there is growing interest in RL-based recipes that apply step-level learning, assigning feedback to intermediate steps rather than final outcomes only~\cite{zha2025rltangoreinforcinggenerator,xu2026incentivizingagenticreasoningllm}. This can be further explored using small open-source models as efficient, privacy-preserving alternatives to proprietary ones.

\section*{Impact Statement}

This paper aims to contribute towards the safe adoption of LLM-based agents for automated network management. A potential social outcome of our work is improved reliability of network operations, contributing to the resilience of Internet-reliant critical services --- network misconfigurations are currently a major cause of Internet outages.


\bibliography{refs}

@misc{yang2024sweagentagentcomputerinterfacesenable,
      title={SWE-agent: Agent-Computer Interfaces Enable Automated Software Engineering}, 
      author={John Yang and Carlos E. Jimenez and Alexander Wettig and Kilian Lieret and Shunyu Yao and Karthik Narasimhan and Ofir Press},
      year={2024},
      eprint={2405.15793},
      archivePrefix={arXiv},
      primaryClass={cs.SE},
      url={https://arxiv.org/abs/2405.15793}, 
}

@misc{jimenez2024swebenchlanguagemodelsresolve,
      title={SWE-bench: Can Language Models Resolve Real-World GitHub Issues?}, 
      author={Carlos E. Jimenez and John Yang and Alexander Wettig and Shunyu Yao and Kexin Pei and Ofir Press and Karthik Narasimhan},
      year={2024},
      eprint={2310.06770},
      archivePrefix={arXiv},
      primaryClass={cs.CL},
      url={https://arxiv.org/abs/2310.06770}, 
}

@misc{vero2025baxbenchllmsgeneratecorrect,
      title={BaxBench: Can LLMs Generate Correct and Secure Backends?}, 
      author={Mark Vero and Niels Mündler and Victor Chibotaru and Veselin Raychev and Maximilian Baader and Nikola Jovanović and Jingxuan He and Martin Vechev},
      year={2025},
      eprint={2502.11844},
      archivePrefix={arXiv},
      primaryClass={cs.CR},
      url={https://arxiv.org/abs/2502.11844}, 
}

@inproceedings{confucius,
author = {Wang, Zhaodong and Lin, Samuel and Yan, Guanqing and Ghorbani, Soudeh and Yu, Minlan and Zhou, Jiawei and Hu, Nathan and Baruah, Lopa and Peters, Sam and Kamath, Srikanth and Yang, Jerry and Zhang, Ying},
title = {Intent-Driven Network Management with Multi-Agent LLMs: The Confucius Framework},
year = {2025},
isbn = {9798400715242},
publisher = {Association for Computing Machinery},
address = {New York, NY, USA},
url = {https://doi.org/10.1145/3718958.3750537},
doi = {10.1145/3718958.3750537},
booktitle = {Proceedings of the ACM SIGCOMM 2025 Conference},
pages = {347–362},
numpages = {16},
location = {S\~{a}o Francisco Convent, Coimbra, Portugal},
series = {SIGCOMM '25}
}

@misc{jain2024livecodebenchholisticcontaminationfree,
      title={LiveCodeBench: Holistic and Contamination Free Evaluation of Large Language Models for Code}, 
      author={Naman Jain and King Han and Alex Gu and Wen-Ding Li and Fanjia Yan and Tianjun Zhang and Sida Wang and Armando Solar-Lezama and Koushik Sen and Ion Stoica},
      year={2024},
      eprint={2403.07974},
      archivePrefix={arXiv},
      primaryClass={cs.SE},
      url={https://arxiv.org/abs/2403.07974}, 
}

@inproceedings {netassistant,
author = {Haopei Wang and Anubhavnidhi Abhashkumar and Changyu Lin and Tianrong Zhang and Xiaoming Gu and Ning Ma and Chang Wu and Songlin Liu and Wei Zhou and Yongbin Dong and Weirong Jiang and Yi Wang},
title = {{NetAssistant}: Dialogue Based Network Diagnosis in Data Center Networks},
booktitle = {21st USENIX Symposium on Networked Systems Design and Implementation (NSDI 24)},
year = {2024},
isbn = {978-1-939133-39-7},
address = {Santa Clara, CA},
pages = {2011--2024},
url = {https://www.usenix.org/conference/nsdi24/presentation/wang-haopei},
publisher = {USENIX Association},
month = apr
}

@inproceedings{bian,
author = {Wang, Chenxu and Zhang, Xumiao and Lu, Runwei and Lin, Xianshang and Zeng, Xuan and Zhang, Xinlei and An, Zhe and Wu, Gongwei and Gao, Jiaqi and Tian, Chen and Chen, Guihai and Liu, Guyue and Liao, Yuhong and Lin, Tao and Cai, Dennis and Zhai, Ennan},
title = {Towards LLM-Based Failure Localization in Production-Scale Networks},
year = {2025},
isbn = {9798400715242},
publisher = {Association for Computing Machinery},
address = {New York, NY, USA},
url = {https://doi.org/10.1145/3718958.3750505},
doi = {10.1145/3718958.3750505},
booktitle = {Proceedings of the ACM SIGCOMM 2025 Conference},
pages = {496–511},
numpages = {16},
location = {S\~{a}o Francisco Convent, Coimbra, Portugal},
series = {SIGCOMM '25}
}

@misc{yao2023reactsynergizingreasoningacting,
      title={ReAct: Synergizing Reasoning and Acting in Language Models}, 
      author={Shunyu Yao and Jeffrey Zhao and Dian Yu and Nan Du and Izhak Shafran and Karthik Narasimhan and Yuan Cao},
      year={2023},
      eprint={2210.03629},
      archivePrefix={arXiv},
      primaryClass={cs.CL},
      url={https://arxiv.org/abs/2210.03629}, 
}

@inproceedings {config2spec,
author = {Rudiger Birkner and Dana Drachsler-Cohen and Laurent Vanbever and Martin Vechev},
title = {{Config2Spec}: Mining Network Specifications from Network Configurations },
booktitle = {17th USENIX Symposium on Networked Systems Design and Implementation (NSDI 20)},
year = {2020},
isbn = {978-1-939133-13-7},
address = {Santa Clara, CA},
pages = {969--984},
url = {https://www.usenix.org/conference/nsdi20/presentation/birkner},
publisher = {USENIX Association},
month = feb
}

@inproceedings {batfish,
author = {Ari Fogel and Stanley Fung and Luis Pedrosa and Meg Walraed-Sullivan and Ramesh Govindan and Ratul Mahajan and Todd Millstein},
title = {A General Approach to Network Configuration Analysis},
booktitle = {12th USENIX Symposium on Networked Systems Design and Implementation (NSDI 15)},
year = {2015},
isbn = {978-1-931971-218},
address = {Oakland, CA},
pages = {469--483},
url = {https://www.usenix.org/conference/nsdi15/technical-sessions/presentation/fogel},
publisher = {USENIX Association},
month = may
}

@article{topologyzoo,
title = "The internet topology zoo",
author = "Simon Knight and Nguyen, \{Hung X.\} and Nickolas Falkner and Rhys Bowden and Matthew Roughan",
note = "Funding Information: The authors would like to acknowledge the support of the Australian Research Council through grants DP0985063 and DP110103505, and two Australian Postgraduate Awards.",
year = "2011",
month = oct,
doi = "10.1109/JSAC.2011.111002",
language = "English",
volume = "29",
pages = "1765--1775",
journal = "IEEE Journal on Selected Areas in Communications",
issn = "0733-8716",
publisher = "IEEE, Institute of Electrical and Electronics Engineers",
number = "9",
}

@misc{wang2025openhandsopenplatformai,
      title={OpenHands: An Open Platform for AI Software Developers as Generalist Agents}, 
      author={Xingyao Wang and Boxuan Li and Yufan Song and Frank F. Xu and Xiangru Tang and Mingchen Zhuge and Jiayi Pan and Yueqi Song and Bowen Li and Jaskirat Singh and Hoang H. Tran and Fuqiang Li and Ren Ma and Mingzhang Zheng and Bill Qian and Yanjun Shao and Niklas Muennighoff and Yizhe Zhang and Binyuan Hui and Junyang Lin and Robert Brennan and Hao Peng and Heng Ji and Graham Neubig},
      year={2025},
      eprint={2407.16741},
      archivePrefix={arXiv},
      primaryClass={cs.SE},
      url={https://arxiv.org/abs/2407.16741}, 
}

@misc{zhang2026agenticcontextengineeringevolving,
      title={Agentic Context Engineering: Evolving Contexts for Self-Improving Language Models}, 
      author={Qizheng Zhang and Changran Hu and Shubhangi Upasani and Boyuan Ma and Fenglu Hong and Vamsidhar Kamanuru and Jay Rainton and Chen Wu and Mengmeng Ji and Hanchen Li and Urmish Thakker and James Zou and Kunle Olukotun},
      year={2026},
      eprint={2510.04618},
      archivePrefix={arXiv},
      primaryClass={cs.LG},
      url={https://arxiv.org/abs/2510.04618}, 
}

@misc{xu2026incentivizingagenticreasoningllm,
      title={Incentivizing Agentic Reasoning in LLM Judges via Tool-Integrated Reinforcement Learning}, 
      author={Ran Xu and Jingjing Chen and Jiayu Ye and Yu Wu and Jun Yan and Carl Yang and Hongkun Yu},
      year={2026},
      eprint={2510.23038},
      archivePrefix={arXiv},
      primaryClass={cs.CL},
      url={https://arxiv.org/abs/2510.23038}, 
}

@misc{zha2025rltangoreinforcinggenerator,
      title={RL Tango: Reinforcing Generator and Verifier Together for Language Reasoning}, 
      author={Kaiwen Zha and Zhengqi Gao and Maohao Shen and Zhang-Wei Hong and Duane S. Boning and Dina Katabi},
      year={2025},
      eprint={2505.15034},
      archivePrefix={arXiv},
      primaryClass={cs.LG},
      url={https://arxiv.org/abs/2505.15034}, 
}

@misc{nika,
      title={A Network Arena for Benchmarking AI Agents on Network Troubleshooting}, 
      author={Zhihao Wang and Alessandro Cornacchia and Alessio Sacco and Franco Galante and Marco Canini and Dingde Jiang},
      year={2025},
      eprint={2512.16381},
      archivePrefix={arXiv},
      primaryClass={cs.NI},
      url={https://arxiv.org/abs/2512.16381}, 
}

@misc{netarena,
      title={NetArena: Dynamic Benchmarks for AI Agents in Network Automation}, 
      author={Yajie Zhou and Jiajun Ruan and Eric S. Wang and Sadjad Fouladi and Francis Y. Yan and Kevin Hsieh and Zaoxing Liu},
      year={2026},
      eprint={2506.03231},
      archivePrefix={arXiv},
      primaryClass={cs.NI},
      url={https://arxiv.org/abs/2506.03231}, 
}

@misc{zheng2023judgingllmasajudgemtbenchchatbot,
      title={Judging LLM-as-a-Judge with MT-Bench and Chatbot Arena}, 
      author={Lianmin Zheng and Wei-Lin Chiang and Ying Sheng and Siyuan Zhuang and Zhanghao Wu and Yonghao Zhuang and Zi Lin and Zhuohan Li and Dacheng Li and Eric P. Xing and Hao Zhang and Joseph E. Gonzalez and Ion Stoica},
      year={2023},
      eprint={2306.05685},
      archivePrefix={arXiv},
      primaryClass={cs.CL},
      url={https://arxiv.org/abs/2306.05685}, 
}

@misc{cornetto,
      title={Benchmarking LLM-Driven Network Configuration Repair}, 
      author={Ioannis Protogeros and Rufat Asadli and Benjamin Hoffman and Laurent Vanbever},
      year={2026},
      eprint={2604.22513},
      archivePrefix={arXiv},
      primaryClass={cs.NI},
      url={https://arxiv.org/abs/2604.22513}, 
}

@misc{zeng2026veriequivbenchequivalencescoregroundtruthfree,
      title={VeriEquivBench: An Equivalence Score for Ground-Truth-Free Evaluation of Formally Verifiable Code}, 
      author={Lingfei Zeng and Fengdi Che and Xuhan Huang and Fei Ye and Xu Xu and Binhang Yuan and Jie Fu},
      year={2026},
      eprint={2510.06296},
      archivePrefix={arXiv},
      primaryClass={cs.PL},
      url={https://arxiv.org/abs/2510.06296}, 
}

@misc{prince2025cloudflare,
  author    = {Prince, Matthew},
  title     = {Cloudflare outage on {November} 18, 2025},
  year      = {2025},
  howpublished = {Cloudflare Blog},
  url       = {https://blog.cloudflare.com/18-november-2025-outage/},
  note      = {Accessed: 2026-04-29}
}

@misc{janardhan2021facebook,
  author       = {Janardhan, Santosh},
  title        = {More details about the {October} 4 outage},
  year         = {2021},
  howpublished = {Engineering at Meta Blog},
  url          = {https://engineering.fb.com/2021/10/05/networking-traffic/outage-details/},
  note         = {Accessed: 2026-04-29}
}

@inproceedings{krentsel-modelfree,
author = {Krentsel, Alexander and Ye, Oliver and Tafoya, Anthony and Ma, Xuqian and Ratnasamy, Sylvia and Shaikh, Anees},
title = {Towards Accessible Model-Free Verification},
year = {2025},
isbn = {9798400722806},
publisher = {Association for Computing Machinery},
address = {New York, NY, USA},
url = {https://doi.org/10.1145/3772356.3772380},
doi = {10.1145/3772356.3772380},
booktitle = {Proceedings of the 24th ACM Workshop on Hot Topics in Networks},
pages = {210–217},
numpages = {8},
location = {UMD Campus, College Park, MD, USA},
series = {HotNets '25}
}

@inproceedings{understanding-misconfigurations,
author = {Mahajan, Ratul and Wetherall, David and Anderson, Tom},
title = {Understanding BGP misconfiguration},
year = {2002},
isbn = {158113570X},
publisher = {Association for Computing Machinery},
address = {New York, NY, USA},
url = {https://doi.org/10.1145/633025.633027},
doi = {10.1145/633025.633027},
booktitle = {Proceedings of the 2002 Conference on Applications, Technologies, Architectures, and Protocols for Computer Communications},
pages = {3–16},
numpages = {14},
location = {Pittsburgh, Pennsylvania, USA},
series = {SIGCOMM '02}
}

@book{Shamimetal2002,
  author    = {Shamim, Faraz and Aziz, Zaheer and Liu, Johnson and Martey, Abe},
  title     = {Troubleshooting IP Routing Protocols},
  series    = {CCIE Professional Development Series},
  publisher = {Cisco Press},
  year      = {2002},
  isbn      = {9780133034684},
  url       = {https://www.ciscopress.com/store/troubleshooting-ip-routing-protocols-ccie-professional-9780133034684}
}

@techreport{rfc4456,
  author      = {Bates, Tony and Chen, Enke and Chandra, Ravi},
  title       = {BGP Route Reflection: An Alternative to Full Mesh Internal BGP (IBGP)},
  howpublished = {Internet Requests for Comments},
  type        = {RFC},
  number      = {4456},
  year        = {2006},
  month       = {April},
  publisher   = {RFC Editor},
  institution = {RFC Editor},
  url         = {https://www.rfc-editor.org/rfc/rfc4456.txt}
}
\bibliographystyle{icml2026}

\newpage
\appendix
\onecolumn
\section{Additional Results}
In this part, we report additional results that complement the main evaluation: the diagnosis and localization scores comparing agentic and monolithic setups (App.~\ref{app:diag-loc}), a per-model decomposition of agent tool-calling trajectories (App.~\ref{app:per-model-trajectory}), and a cost analysis of the agentic pipeline against monolithic baselines (App.~\ref{app:cost-analysis}).

\subsection{Diagnosis and Localization}
\label{app:diag-loc}

\begin{figure*}[h]
    \centering
    \begin{subfigure}[h]{0.49\textwidth}
        \centering
        \includegraphics[width=\textwidth]{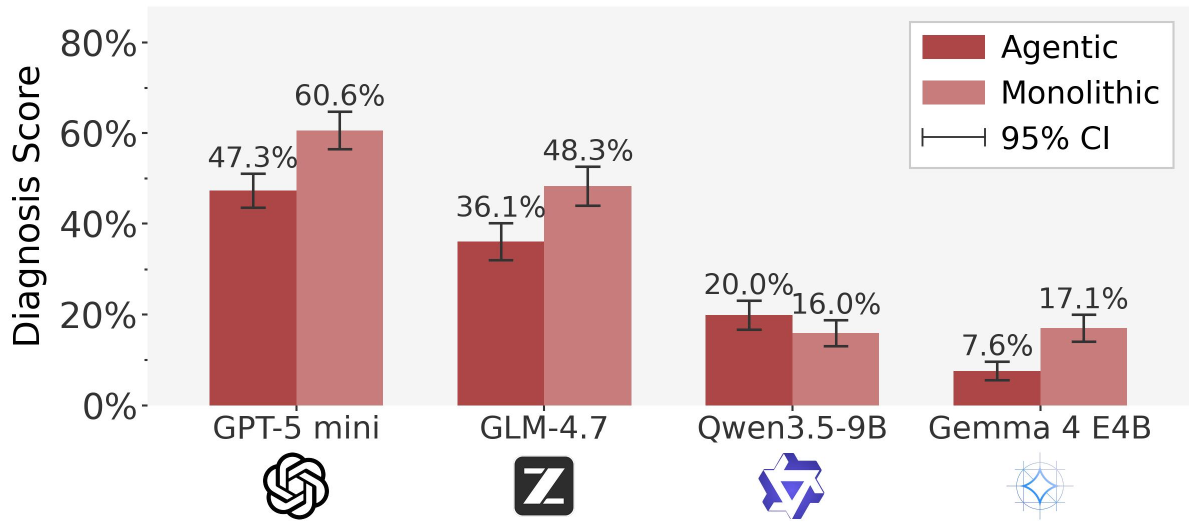}
        \caption{Diagnosis ($\uparrow$ \textit{is better})}
        \label{fig:baseline_vs_agentic_diagnosis}
    \end{subfigure}%
    \hfill%
    \begin{subfigure}[h]{0.49\textwidth}
        \centering
        \includegraphics[width=\textwidth]{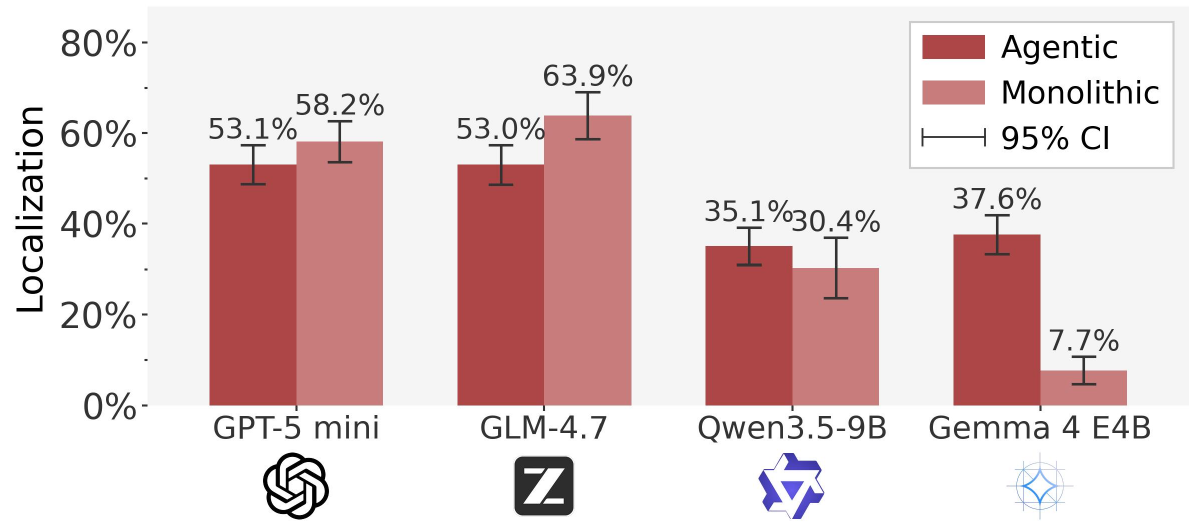}
        \caption{Localization ($\uparrow$ \textit{is better})}
        \label{fig:baseline_vs_agentic_localization}
    \end{subfigure}%
    \caption{Comparison of agentic and monolithic LLMs in terms of diagnosis score and root-cause localization performance (in \%).}
    \label{fig:baseline_v_agentic_diag_localization}
\end{figure*}

\begin{figure*}[h]
    \centering
    \begin{subfigure}[h]{0.49\textwidth}
        \centering
        \includegraphics[width=\textwidth]{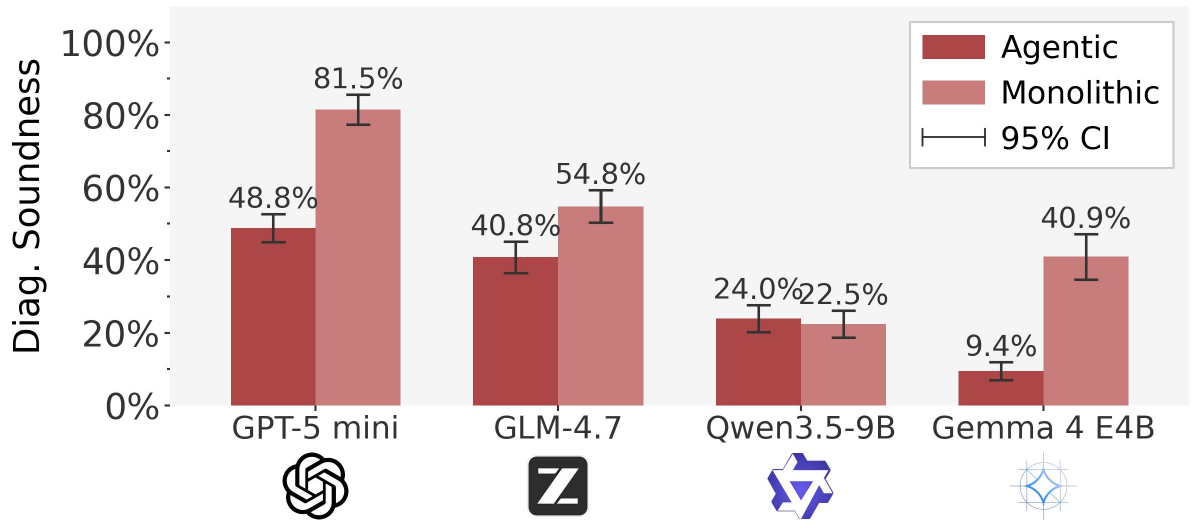}
        \caption{Diagnosis Soundness ($\uparrow$ \textit{is better})}
        \label{fig:baseline_vs_agentic_soundness}
    \end{subfigure}%
    \hfill%
    \begin{subfigure}[h]{0.49\textwidth}
        \centering
        \includegraphics[width=\textwidth]{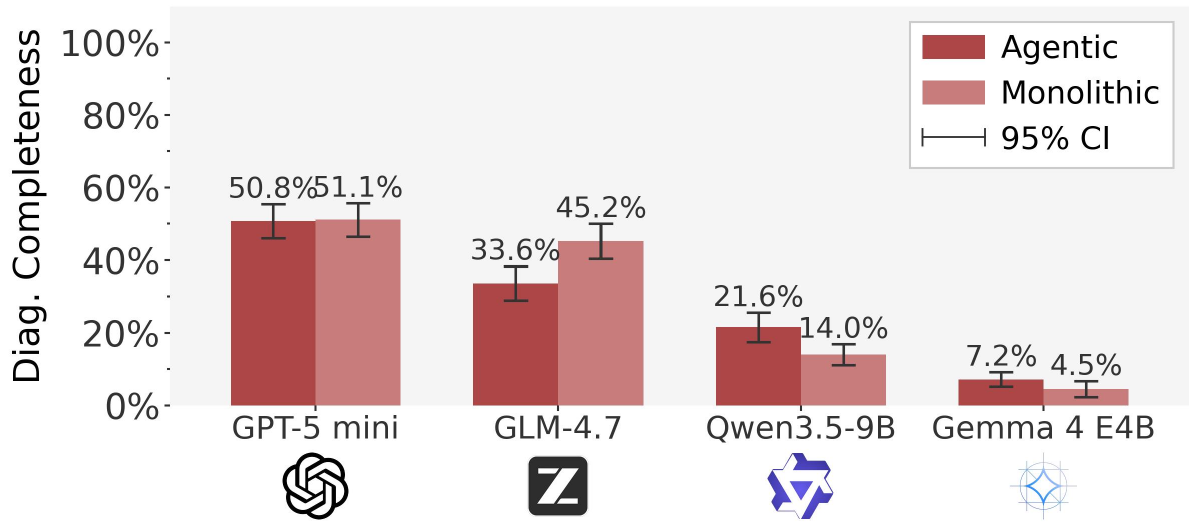}
        \caption{Diagnosis Completeness ($\uparrow$ \textit{is better})}
        \label{fig:baseline_vs_agentic_completeness}
    \end{subfigure}%
    \caption{Comparison of agentic and monolithic LLMs in terms of diagnosis soundness and completeness performance (in \%).}
    \label{fig:baseline_v_agentic_soundness_completeness}
\end{figure*}

Beyond reporting the fix score and regression rate for the agents against monolithic baselines (Sec.~\ref{sec:eval-results}), we present additional metrics in this part: the diagnosis and localization scores (Fig.~\ref{fig:baseline_v_agentic_diag_localization}). Following \textsc{Cornetto}, beyond editing configuration files, we require the agents to provide a textual diagnosis of their solution and to list the files they deem problematic. We report diagnosis scores using an LLM-as-a-Judge method~\cite{zheng2023judgingllmasajudgemtbenchchatbot}, in which three LLM judges receive the ground-truth problem setting and evaluate how well the agents identify the \textit{misconfigurations} (see App.~\ref{app:pipeline-configs} for details). We compute the localization as a classification score, measuring how well the agents identify the \textit{misconfigured files}.

Interestingly, while the agentic setup generally outperforms the monolithic baselines in other metrics, agents tend to achieve lower diagnosis scores. We link this with the diagnosis soundness and completeness patterns shown in Fig.~\ref{fig:baseline_v_agentic_soundness_completeness}. The soundness term measures the proportion of diagnosis claims that are correct (i.e., precision), whereas completeness stands for the proportion of true faults that were actually identified (i.e., recall). We observe that agents appear to adopt a more aggressive behaviour in identifying true positives, yielding comparable recall, but at the cost of lower precision relative to monolithic models. The higher rate of false positives, potentially arising from hallucinations, introduces noise into the agents' overall fault diagnoses, ultimately reducing their diagnostic accuracy.


\subsection{Agent Trajectory}
\label{app:per-model-trajectory}

\begin{figure}[h]
\centering
\includegraphics[width=0.49\linewidth]{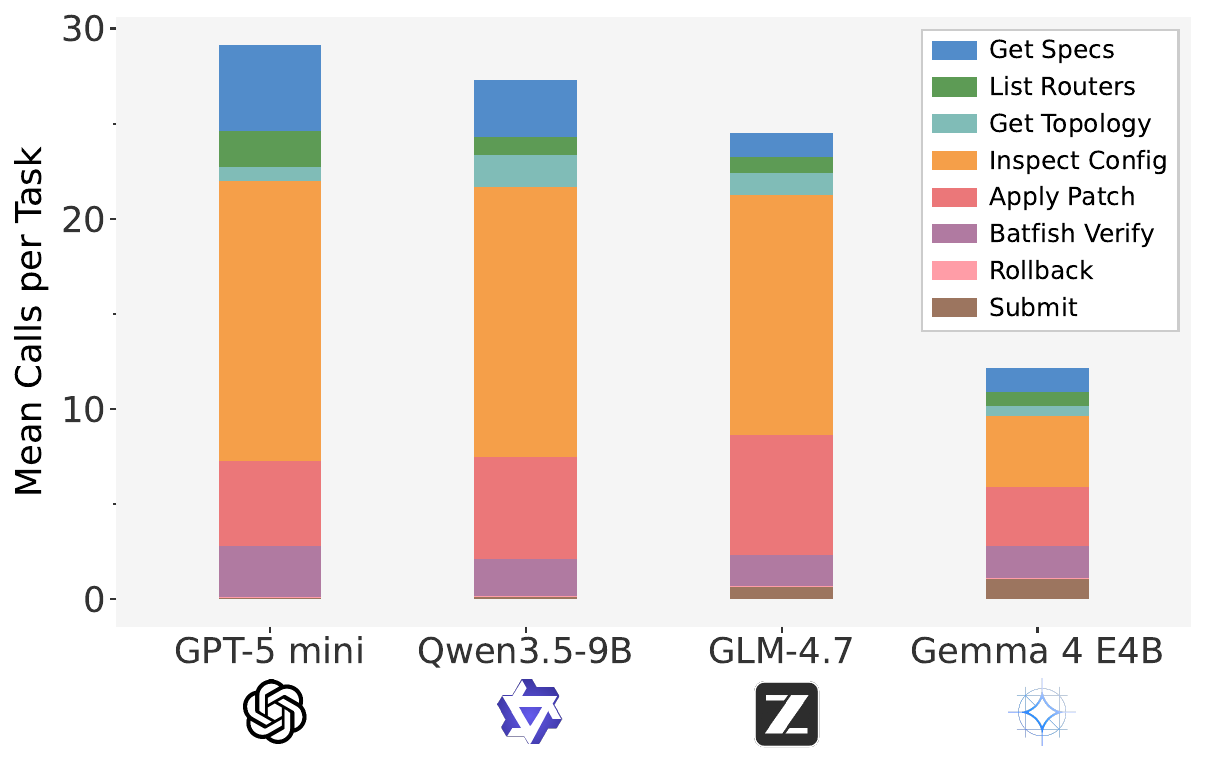}
\captionof{figure}{The frequency of the tools called by each model in a task on average.}
\label{fig:model-tool-calling-trajectory}
\end{figure}

In Figure~\ref{fig:model-tool-calling-trajectory}, we decompose the per-step agent trajectory, aggregated over all models, into separate views for each agent. Agents generally tend to exhaust all the allocated maximum step budgets in a single task, although \textsc{Gemma 4 E4B} uses roughly half of its budget only, given its limited capacity to retain reasonable performance over a wider interaction horizon---
following a very similar tool-calling distribution as in the aggregated case (Fig.~\ref{fig:tool_calling_counts_avg}). Notably, agents spend most of the iterations inspecting file contents. On average, they utilize the verifier feedback 2-3 times per task, opening a potential discussion for verification mechanisms with more powerful feedback signals.
In future work, we believe that further analysis of agent behavior \textemdash~more specifically, patterns in the order of tool usage across interaction trajectories \textemdash~can yield valuable insights into agentic reasoning.

\subsection{Cost-Performance Trade-off}
\label{app:cost-analysis}

\begin{figure*}[h]
    \centering
    \begin{subfigure}[h]{0.49\textwidth}
        \centering
        \includegraphics[width=\textwidth]{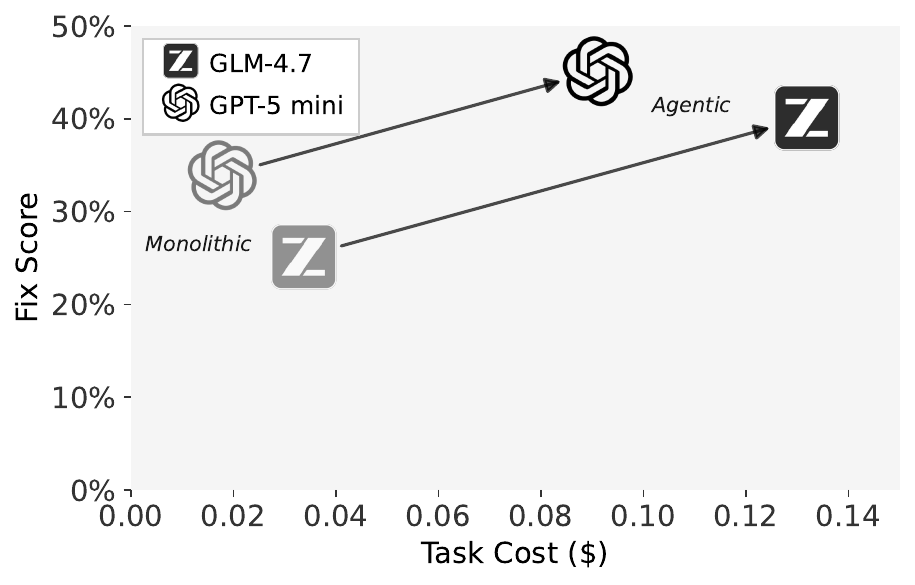}
        \caption{Fix Score $(\uparrow)$ \& Cost $(\downarrow)$}
        \label{fig:cost_v_fix_score}
    \end{subfigure}%
    \hfill%
    \begin{subfigure}[h]{0.49\textwidth}
        \centering
        \includegraphics[width=\textwidth]{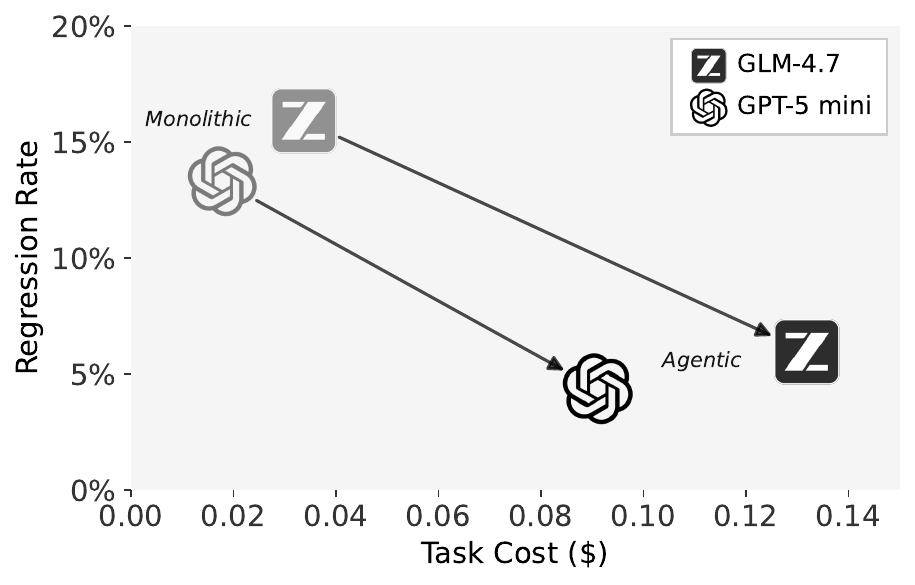}
        \caption{Regression Rate $(\downarrow)$ \& Cost $(\downarrow)$}
        \label{fig:cost_v_regression}
    \end{subfigure}%
    \caption{Comparison of agentic and monolithic LLMs in terms of fix score and regression rate, against the per task cost (in \$). Model identifiers, distinguished by the original and faded logos, stand for the agentic and monolithic setups, respectively.}
    \label{fig:cost_v_fix_regression}
\end{figure*}

The superior performance of the agentic pipeline comes at a higher inference cost. As shown in Fig.~\ref{fig:cost_v_fix_regression}, with a budget of 30 maximum steps, \textsc{GPT-5 mini} spends \$0.091 per task against \$0.018 for the monolithic baseline (roughly a 5$\times$ increase). Similarly, the average per-task cost of running \textsc{GLM-4.7} in the agentic setup increases by around 4$\times$ from \$0.034 to \$0.132. Therefore, as previously emphasized in Sec.~\ref{sec:eval-results}, this motivates identifying the optimal step budget that recovers most of the benefit as key to controlling the cost of agentic repair in practice.

\paragraph{Example cost-step analysis} In Table~\ref{tab:cost-tradeoff}, we compare the monolithic baseline to the agentic setup at different iteration budgets and pipeline modes. With iterative verifier feedback enabled, at just 10 steps, \textsc{GPT-5 mini} already reaches a 28.5\% fix score (vs. 33.9\% monolithic), while cutting the regression rate from 13.3\% to 2.0\% \textemdash~at a cost of only \$0.023, barely above the monolithic baseline. By 20 steps, the agentic pipeline surpasses the baseline by 10\% with a fix score of 42.7\% and holds regression below 1\%, for roughly \$0.050 per task. 
With feedback mode, going from 20 to 30 steps yields a very similar cost with a marginal drop in performance metrics, which might be due to inherent randomness in model responses and also changes in the tool-calling trajectory. This points to diminishing returns beyond 20 steps, as the most noticeable gains occur in the first 10-20 iterations. However, this observation is based on a three-tier step budget analysis and a single model. The precise inflection point may shift with more granular step increments, different models, or task-specific tools.

Lastly, with verifier feedback disabled, increasing the step budget to 30 iterations almost doubles the cost, with the fix score improving to 45.2\% but the regression rate decreases by 3.1 percentage points. Comparing the two pipeline variants at 30 steps reveals the role of the verifier: when verifier feedback is disabled, the agent can only utilize the context retrieval tools, which return more information-rich context about the network state (e.g., configuration files, topology, specification). These components consume additional tokens at each iteration and thus compound the total cost.

\begin{table}[t]
\centering
\caption{Cost–performance trade-off for \textsc{GPT-5 mini} across pipeline configurations. Against the monolithic baseline, we demonstrate the performance of the agentic setup with and without access to the verifier feedback as a tool.}
\label{tab:cost-tradeoff}
\begin{tabular}{llccc}
\toprule
\textbf{Pipeline} & \textbf{Steps} & \textbf{Fix Score (\%)} & \textbf{Regression (\%)} & \textbf{Cost (\$)} \\
\midrule
Monolithic                &  1 & 33.9 & 13.3 & 0.018 \\
\midrule
Agentic w/o feedback      & 30 & 45.2 &  4.4 & 0.091 \\
\midrule
Agentic w/ feedback       & 10 & 28.5 &  2.0 & 0.023 \\
Agentic w/ feedback       & 20 & 42.7 &  0.9 & 0.050 \\
Agentic w/ feedback 
& 30 & 40.8 &  1.3 & 0.049 \\
\bottomrule
\end{tabular}
\end{table}

\section{Experimental Details}
To support the results in Sec.~\ref{sec:eval-results}, we detail the design choices underlying our experiments: the statistics of the \textsc{Cornetto} benchmark test set (App.~\ref{app:dataset-stats}), the model selection spanning a capability spectrum (App.~\ref{app:model-selection}), and the pipeline configurations covering LLM-as-a-Judge diagnosis, interaction-history management, and context sampling (App.~\ref{app:pipeline-configs}). Finally, we include a comprehensive list of benchmark misconfigurations in App.~\ref{app:faults} for completeness.

\subsection{Dataset Statistics}
\label{app:dataset-stats}

To directly compare our setup to the monolithic baseline, we evaluate the agents on the \textsc{Cornetto} test set, which is modeled after frequently observed real-world outages. As mentioned earlier, it comprises 231 misconfiguration scenarios across topologies from the Topology Zoo, spanning 27 fault types with up to 8 simultaneous faults per scenario. Fault injection mechanisms are minimal yet highly disruptive in impact, requiring agents to have a solid domain knowledge to reason about accurately locating and repairing the faults throughout the iterative process.

\begin{table}[h]
\centering
\captionof{table}{\textsc{Cornetto} dataset statistics~\cite{cornetto}.}
\label{tab:dataset}
\small
\begin{tabular}{llrr}
\toprule
& \textbf{Metric} & \textbf{Mean} & \textbf{Max} \\
\midrule
Topology     & Nodes (\#)            & 86.4  & 754    \\
             & Config lines          & 16.1K & 200.0K \\
             & Data plane predicates & 12.7K & 598.0K \\
\addlinespace
Fault impact & Routers affected      & 5.5   & 20     \\
\addlinespace
Impact (\%)  & LoC\% changed         & 0.44  & 5.93   \\
             & Routes\% changed      & 6.34  & 57.3   \\
             & Predicates\% changed  & 6.82  & 49.8   \\
\bottomrule
\end{tabular}
\end{table}


\subsection{Model Selection}
\label{app:model-selection}
To show the feasibility of our minimal agentic setup over monolithic architectures, we first conduct a focused comparison using four models spanning a range of capabilities: \textsc{GPT-5 mini} and \textsc{GLM-4.7} as strong proprietary and closed-source representatives, alongside \textsc{Qwen3.5-9B} and \textsc{Gemma 4 E4B} as smaller open-source alternatives. Rather than aiming for exhaustive model coverage, this selection allows us to test whether the agentic benefit holds across a capability spectrum  \textemdash~from models that already perform reasonably on \textsc{Cornetto} to those that struggle in the monolithic setting. As we show in Sec.~\ref{sec:eval-results}, the pattern is consistent: agentic scaffolding improves fix scores and reduces regressions for all selected models, with disproportionately large gains for open-source models of lower capacities.
Lastly, the agents are configured with the following hyperparameters across all experiments: (1) \texttt{max}\_\texttt{tokens} of $40,000$, (2) \texttt{top-}\textit{\texttt{p}} of 0.95, and (3) \texttt{temperature} of 0.7.

\subsection{Pipeline Configurations}
\label{app:pipeline-configs}

Across our agentic experiments, we adhere to the following pipeline configurations.

\paragraph{LLM-as-a-Judge diagnosis} Beyond fix score and regression rate, which capture direct repair performance, we also evaluate models' diagnostic and root-cause identification abilities. For this, we adopt an LLM-as-a-Judge~\cite{zheng2023judgingllmasajudgemtbenchchatbot} approach in which three LLMs (\textsc{GPT-5.1}, \textsc{Claude 4.5 Opus}, and \textsc{Gemini 2.5 Pro}) independently assess the soundness and completeness of each model's solution trajectory. We follow a multi-judge evaluation mechanism to avoid potential biases. More importantly, each judge receives the ground truth misconfiguration (i.e., the list of injected faults), meaning that agent diagnoses are rewarded only when consistent with ground truth.

\paragraph{Dynamic interaction history} As in prior work~\cite{yang2024sweagentagentcomputerinterfacesenable}, in our experiments, we grant agents access to the full running chat history at each step. In general, each iteration produces a thought-action-observation triple that is appended to the conversation. As the chat history grows, older tool outputs, particularly verbose configuration dumps and specification lists, can consume a significant portion of the context window. Following the windowing strategy from SWE-agent, we truncate observations older than the most recent 5 steps, replacing them with a short placeholder while preserving the agent's own reasoning (i.e., thought and action messages) in full. This allows the model to recall its prior decisions without accumulating additional token cost of retaining large tool outputs that it has already processed. However, for low capacity models, it might be harder to retain the accumulating interaction history, as we see in the case of \textsc{Qwen3.5-9B} with marginally increasing safety regressions across longer interaction cycles (Fig.~\ref{fig:ablation_max_steps}).

\paragraph{Context sampling} Baseline experiments from \textsc{Cornetto} follow two context management strategies to handle information overflow across models with varying context windows: \textit{oracle} mode, which provides only the configuration files that were actually modified by the fault, and \textit{random} mode, which fills the available context window with a random subset of configurations (with a fixed randomness seed for all experiments), constrained by each model's own context window limits. Therefore, it is technically possible that a model, with a large enough context window, will have access to all configuration files available.
In our agentic setup, we adopt the random sampling strategy, in which the agent can access all selected configuration files through its retrieval tools. However, unlike the monolithic baseline, which must fit all context into a single prompt, the agent retrieves configurations on demand, inspecting only those it deems relevant to the diagnosed fault. This effectively decouples context availability from context window pressure, as configurations are loaded incrementally rather than all at once.

\newpage
\subsection{Fault Types}
\label{app:faults}

The benchmark contains various misconfigurations that reflect common human errors encountered in the wild. Faults are grounded in past incidents~\cite{janardhan2021facebook}, Request for Comments (RFCs)~\cite{rfc4456}, and troubleshooting manuals~\cite{Shamimetal2002}.

\begin{center}
    
\centering
\scriptsize 
\captionof{table}{Comprehensive fault catalog listing the protocols affected, the nature of the misconfiguration (Summary), and the resulting impact on the network (Expected Effect).~\cite{cornetto}}
\label{tab:fault_catalog}
\begin{tabularx}{\textwidth}{l p{0.35\textwidth} X} 
\toprule
\textbf{Protocol / Type} & \textbf{Summary} & \textbf{Expected Effect} \\
\midrule

\textbf{BGP} 
 & eBGP neighbor configured with incorrect remote AS 
 & eBGP session reset due to ASN mismatch, cutting off inter-AS route exchange \\
 
 & Administratively shut down a BGP neighbor 
 & BGP Peering is disabled, withdrawing all prefixes learnt via the neighbor \\

& Node configured with incorrect local ASN 
 & Misaligned local ASN breaks iBGP/eBGP sessions and splits the AS control plane \\
 
 & Force invalid next-hop on eBGP advertisements 
 & Outbound policy rewrites next-hop to an unreachable address, causing downstream traffic blackholes \\
 
 & Remove next-hop-self from RR $\rightarrow$ client iBGP session 
 & iBGP routes advertised to clients retain original eBGP next-hop, which may be unreachable from clients causing traffic blackholes \\
 
 & Withdraw a BGP network statement from the process 
 & Prefix is no longer originated, withdrawing reachability from downstream peers \\
 
 & Remove outbound route-map from eBGP neighbor 
 & Export policy no longer enforced, allowing infrastructure routes (loopbacks, P2P) and unintended prefixes to leak to external peers \\
 
 & Swap inbound and outbound route-maps on a neighbor 
 & Inbound filters begin applying outbound and vice versa, breaking intended import/export policy \\
 
 & Leak router loopback by stripping export/import policies 
 & ASBR originates its loopback /32 into eBGP and the peer accepts it because inbound filtering was removed \\
 
 & Break RR sessions to orphan clients 
 & iBGP sessions removed between RR and up to 5 (exclusive) clients, orphaning them from iBGP reachability \\
 
 & Duplicate cluster-id across route reflectors and isolate clients on one RR 
 & Conflicting cluster-ids cause route reflectors to drop one another's updates, stranding clients that now depend on the misconfigured RR (cf. RFC 4456, Sec. 8) \\
\midrule

\textbf{OSPF} 
 & OSPF interface cost set to extreme value 
 & Artificially high OSPF cost diverts traffic away from the link based on alternate SPF paths \\
 
 & Disable OSPF adjacency on a link 
 & Removing the link from OSPF prevents adjacency formation and withdraws LSAs learned across it \\
 
 & Node missing OSPF area membership 
 & Router withdraws from all OSPF areas, tearing down adjacencies and LSAs \\
 
 & Assign duplicate OSPF router-ID to multiple routers 
 & OSPF adjacencies fail or LSAs rejected due to router-ID collision, fragmenting OSPF domain and blackholing traffic \\
\midrule

\textbf{IS-IS} 
 & Disable IS-IS on an intra-AS link 
 & Removing the link from IS-IS prevents adjacency formation and withdraws LSPs learned across it \\
 
 & Demote a Level-1-2 IS-IS router to Level-1 
 & Reduces inter-area reachability by removing a backbone-capable router, risking L2 partitioning \\
 
 & Assign router to wrong IS-IS area 
 & Router in wrong area cannot form L1 adjacencies with its physical neighbors; causes partition of L1 domain and reachability loss \\
\midrule

\textbf{Addressing} 
 & Duplicate loopback IPv4 addresses 
 & Two routers share the same loopback, risking routing loops and control-plane instability \\
 
 & Link interfaces disagree on prefix length 
 & One side of a point-to-point link uses a mismatched subnet mask, preventing adjacency formation \\
 
 & Link interfaces reside in different subnets 
 & Interfaces on a point-to-point link move to disjoint IPv4 subnets, breaking adjacency formation \\
\midrule

\textbf{Device} 
 & Remove supporting static route for advertised prefix 
 & Advertised network disappears once the backing static route is withdrawn, causing a control-plane withdraw \\
\midrule

\textbf{Policy} 
 & Remove permit entry from prefix-list 
 & Prefix-list no longer matches intended prefixes, causing route filtering to block previously allowed routes \\
 
 & Convert BGP route-map permit clause into deny 
 & Previously exported prefixes are now filtered, withdrawing routes from neighbors \\
 
 & Lower BGP local-preference on inbound policy 
 & Reduced local-preference makes an alternate egress the best path for affected prefixes \\
\midrule

\textbf{Redistribution} 
 & Drop BGP $\rightarrow$ OSPF redistribution on an ASBR 
 & Internal OSPF loses external reachability because Type-5 LSAs are never originated \\
 
\midrule

\textbf{Security} 
 & Insert implicit deny at top of interface ACL 
 & Ingress traffic on the protected interface is dropped before policy permits, breaking connectivity \\
 
 & Insert implicit deny at top of outbound interface ACL 
 & Egress traffic on the protected interface is dropped before policy permits, breaking connectivity \\

\bottomrule
\end{tabularx}
\end{center}

\newpage
\section{Prompts}
\subsection{Main Agentic Prompts}
\label{app:agentic-input-prompt}

\begin{tcolorbox}[title=Agent System Prompt, colback=gray!5, colframe=black, colbacktitle=black, coltitle=white, fonttitle=\bfseries]
\ttfamily\small
A network fault has been detected. Please diagnose and repair the misconfiguration.

\medskip
Start by examining the violated specifications to understand what is broken, then inspect the relevant router configurations, apply fixes, and verify.

\medskip
You have a budget of \textcolor{blue}{\{max\_steps\}} tool calls for this task.
\end{tcolorbox}

\begin{tcolorbox}[title=Agent Input Prompt, colback=gray!5, colframe=black, colbacktitle=black, coltitle=white, fonttitle=\bfseries, breakable]
\ttfamily\small
You are an expert network engineer agent. Your task is to diagnose and repair network misconfigurations by iteratively inspecting configurations, applying targeted fixes, and verifying correctness using formal data-plane analysis.

\medskip
\textbf{\ttfamily\normalsize Available Tools}
\medskip

You interact with the network environment through tool calls. Each turn, you should output your reasoning (Thought), then a single tool call (Action).

\medskip
\textcolor{blue}{\{tool\_descriptions\}}
\medskip

\textbf{\ttfamily\normalsize Tool Call Format}
\medskip

These tools are NOT native function calls. You must invoke them using plain text in the EXACT format shown below. Do not use any other tool-calling syntax, function-calling format, or internal protocol. Simply output the following three lines as plain text:

\medskip
Thought: <your reasoning about what to do next>\\
Action: <tool\_name>\\
Action Input: <JSON object with parameters, or \{\} for no-parameter tools>

\medskip
Examples:

\medskip
\hspace{1em}Thought: I need to see which routers are in the network.\\
\hspace{1em}Action: list\_routers\\
\hspace{1em}Action Input: \{\}

\medskip
\hspace{1em}Thought: Let me inspect the BGP configuration of router 0\_as65003\_0.cfg.\\
\hspace{1em}Action: inspect\_config\\
\hspace{1em}Action Input: \{"router\_name": "0\_as65003\_0.cfg"\}

\medskip
\hspace{1em}Thought: I found the misconfigured subnet mask. Let me fix it.\\
\hspace{1em}Action: apply\_patch\\
\hspace{1em}Action Input: \{"router\_name": "0\_as65003\_0.cfg", "search": " ip address 10.0.0.1 255.255.255.0", "replace": " ip address 10.0.0.1 255.255.255.252"\}

\medskip
\hspace{1em}Thought: I've applied my patches. Let me verify the network state.\\
\hspace{1em}Action: verify\\
\hspace{1em}Action Input: \{\}

\medskip
\hspace{1em}Thought: All specs are fixed with no regressions. I'll submit.\\
\hspace{1em}Action: submit\\
\hspace{1em}Action Input: \{\}

\medskip
\textbf{\ttfamily\normalsize Strategy}
\medskip

1. Start by examining the violated specifications to understand what is broken.\\
2. Inspect relevant router configurations to diagnose root causes.\\
3. Apply targeted patches to fix the identified issues.\\
4. Run verification to check if specs are restored and no regressions occurred.\\
5. If issues remain, analyze the verification feedback and iterate.\\
6. If a patch introduces regressions, rollback and try a different approach.\\
7. Submit when satisfied or when you've exhausted reasonable repair attempts.

\medskip
\textbf{\ttfamily\normalsize Important Rules}
\medskip
\begin{itemize}[leftmargin=1.5em, itemsep=0pt]
\item Be precise with search blocks: they must match the config EXACTLY (whitespace, indentation, etc.).
\item Prefer small, targeted patches over large rewrites.
\item Always verify after applying patches before submitting.
\item If verification shows regressions, consider rolling back.
\item You have a limited budget of \textcolor{blue}{\{max\_steps\}} tool calls. Use them wisely.
\end{itemize}
\end{tcolorbox}

\subsection{Tool Stack}
\label{app:tools-descriptions}
\begin{tcolorbox}[title=Agent Tool Definitions, colback=gray!5, colframe=black, colbacktitle=black, coltitle=white, fonttitle=\bfseries]
\ttfamily\small
\textbf{list\_routers} --- List all router configuration filenames available in the network. \textit{Returns:} A list of router filenames.
\medskip

\textbf{inspect\_config} --- Retrieve the full configuration text of a specific router. Use this to examine a router before deciding what to fix. \textit{Returns:} The full configuration text of the specified router.
\medskip

\textbf{get\_violated\_specs} --- Get the list of currently violated network specifications. These describe the gap between intended and actual behavior. \textit{Returns:} Violated specifications with type, source, destination, and status.
\medskip

\textbf{get\_topology} --- Get the network topology (nodes and links). \textit{Returns:} The network topology as a JSON structure.
\medskip

\textbf{apply\_patch} --- Apply a search-and-replace edit to a router configuration. The search block must appear exactly once in the config. \textit{Returns:} Success or failure message with details.
\medskip

\textbf{verify} --- Run formal data-plane verification on the current network state. Returns which specs are fixed, which remain broken, and any regressions. \textit{Returns:} Verification report with fix\_score, regression\_rate, and per-predicate classification.
\medskip

\textbf{rollback} --- Undo ALL patches applied since the last successful verification. Restores configs to the last verified-safe state. \textit{Returns:} Confirmation that configs have been rolled back.
\medskip

\textbf{submit} --- Submit the current configuration as the final solution. Call when satisfied with verification results. \textit{Returns:} Confirmation that the solution has been submitted.
\end{tcolorbox}

\newpage

\subsection{Prefilled Context Prompt}
\begin{tcolorbox}[title=Prefilled Input Prompt, colback=gray!5, colframe=black, colbacktitle=black, coltitle=white, fonttitle=\bfseries, breakable]
\ttfamily\small
A network fault has occurred, possibly due to human error, causing discrepancies in at least one of the configuration files. The specifications that govern the network behavior have changed as a result.
\medskip

\textbf{Task Requirements:}\\
1. Identify which routers must be modified to restore the correct forwarding behavior.\\
2. Generate the necessary modifications for each router configuration.\\
3. Provide your solution as explicit search-and-replace instructions for each router.
\medskip

\textbf{Expected Output Format:}\\
Output must be valid YAML only, with the following top-level keys: \texttt{routers}, then \texttt{metadata}, then \texttt{replacements}.
\medskip

\textbf{Specification Semantics:}\\
Specifications are formatted as CSV lines with columns:\\
\texttt{Type, Source\_Node, Destination\_Prefix, waypoint\_node, num\_routes, Status}
\medskip

Status values:\\
--- \texttt{broken\_removed}: behavior IS required but CURRENTLY MISSING. Restore it.\\
--- \texttt{broken\_added}: behavior IS NOT required but CURRENTLY PRESENT. Remove it.
\medskip

Invariant types: \texttt{reachability}, \texttt{waypointing}, \texttt{isolation}, \texttt{load\_balancing}.
\medskip

\textbf{Network specifications:} \textcolor{blue}{\{preds\_text\}}
\medskip

\textbf{Current network topology:} \textcolor{blue}{\{topology\_text\}}
\medskip

\textbf{Router configurations:}\\
--- START OF \textcolor{blue}{\{filename\}} ---\\
\textcolor{blue}{\{config\}}\\
--- END OF \textcolor{blue}{\{filename\}} ---\\
\textit{(repeated for each router)}
\medskip

All network context has been provided above. The retrieval tools (list\_routers, inspect\_config, get\_violated\_specs, get\_topology) are disabled --- use the information above instead.

\medskip
Proceed directly to diagnosing the issue and applying patches using the Thought/Action/Action Input format.

\medskip
You have a budget of \textcolor{blue}{\{max\_steps\}} tool calls for this task.

\medskip
Begin now.
\end{tcolorbox}

\end{document}